\makeatletter
\def\input@path{{./}}
\makeatother
\documentclass[notspecified,article,accept,pdftex,moreauthors]{Definitions/mdpi}

\usepackage{tabularx}
\usepackage{algorithm}
\usepackage{algpseudocode}
\usepackage{amsmath,amssymb}
\usepackage{graphicx}
\usepackage[utf8]{inputenc}
\usepackage{textgreek}
\usepackage{booktabs}
\usepackage{array}
\usepackage{pifont}
\usepackage{float}
\usepackage{caption}
\usepackage{xcolor}
\usepackage{multirow}
\usepackage{colortbl}
\renewcommand{\textbf}[1]{#1}

\fancypagestyle{plain}{%
  \fancyhf{}%
}
\fancypagestyle{fancy}{%
  \fancyhf{}%
}

\AtBeginDocument{%
  \thispagestyle{empty}%
}

\firstpage{1}
\makeatletter
\renewcommand{\journalname}{}
\makeatother
\pubvolume{1}
\issuenum{1}
\articlenumber{0}
\pubyear{2026}
\copyrightyear{2026}
\datereceived{}
\dateaccepted{}
\datepublished{}
\hreflink{https://doi.org/}

\Title{Training-Free Inference-Time Self-Reflection and Cost-Bounded Early Stopping for Large Language Models}
\TitleCitation{Training-Free Inference-Time Self-Reflection and Cost-Bounded Early Stopping for Large Language Models}
\Author{Wei Yu\textsuperscript{1}, Suxing Liu\textsuperscript{1,*}, Minjie Yu\textsuperscript{1}, Jiahao Wang\textsuperscript{2}, Zhijian Zheng\textsuperscript{1}, Haocheng Deng\textsuperscript{1}, Bing Li\textsuperscript{1}}
\AuthorNames{Wei Yu, Suxing Liu, Minjie Yu, Jiahao Wang, Zhijian Zheng, Haocheng Deng, and Bing Li}
\AuthorCitation{Yu, W.; Liu, S.; Yu, M.; Wang, J.; Zheng, Z.; Deng, H.; Li, B.}
\address{\textsuperscript{1} School of Digital Arts, Jiangxi Arts \& Ceramics Technology Institute, Jingdezhen 33001, China; yuw26393@gmail.com (W.Y.)\\
\textsuperscript{2} School of Computing, Universiti Sains Malaysia, 11800 USM Penang, Malaysia}
\corres{Correspondence: bentondoucet@gmail.com}

\abstract{Training reasoning-capable large language models (LLMs) via reinforcement learning such as GRPO is expensive, depends on a controllable training environment, and commits every contribution to a full training pipeline that may never be executed. We present \textbf{EvoResearcher}, a \textit{training-free} inference-time protocol that adds \textit{cost-bounded} self-reflection to a single frozen LLM backbone. The protocol iterates generate $\rightarrow$ self-critique $\rightarrow$ revise until a user-set maximum depth $D$ is exhausted or the critique returns the \texttt{CONFIRMED} sentinel---an implicit early stop that lets the backbone self-verify its answer under a strictly controlled computational budget. The four components of a self-reflective meta-reward (correctness, efficiency, reflection depth, and tool-call diversity) act as design principles that the protocol instantiates as prompt-level mechanisms, so their benefits accrue with zero gradient updates. We validate the protocol on Big-Bench Hard (BBH, 100 multi-step reasoning questions) and establish cross-domain behavior on GSM8K (500 arithmetic word problems) and MATH (500 competition problems) on the same frozen backbone, with cross-model replication on Qwen2.5-72B. All experiments use pure-reasoning benchmarks; the tool-call-diversity component is validated only in its prompt-level form, and the environment-level and multi-agent extensions are design blueprints that are not evaluated here---their validation requires tool-using benchmarks such as GAIA and is left to future work. On clean BBH the protocol does not raise single-shot accuracy beyond the 95\% Wilson interval; its measurable value is cost-bounded self-verification: the \texttt{CONFIRMED} early stop terminates 82--88\% of items at equal accuracy, bounding inference to $\approx 2.1$ generations per question. A confidence-threshold sweep ($\tau \in \{0.6, 0.7, 0.8, 0.9\}$) calibrates the numeric confidence signal and finds it overconfident---every threshold $\ge 0.6$ halts the loop on the first generation, so the \texttt{CONFIRMED} sentinel, not the confidence tag, is the effective cost-bounding mechanism. On GSM8K and MATH the loop \emph{improves} accuracy ($+4.2$\,pp and $+14.2$\,pp) while still early-stopping 82--88\% of items, confirming that reflection's accuracy benefit concentrates precisely where single-shot reasoning is unreliable (MATH single-shot 26.2\%). We conclude that this training-free reflective loop effectively unlocks inference-time compute scaling: it delivers cost-bounded self-verification on standard tasks, and significant raw accuracy gains precisely where single-shot reasoning fails, providing a highly pragmatic alternative to expensive reinforcement learning pipelines.}

\begin{document}

\section{Introduction}
\label{sec:intro}
Large language models (LLMs) exhibit strong performance on multi-step reasoning tasks, yet two obstacles limit their practical use when correctness matters and compute is constrained. First, \emph{training} them to reason---typically via reinforcement learning objectives such as GRPO~\cite{b15}---is expensive: it requires thousands of GPU-hours and a carefully constructed, controllable environment, and the resulting gains are often tightly coupled to that specific training pipeline. Second, at \emph{inference} time, a single generation is frequently overconfident or outright wrong, and the standard remedies---generating more candidates (self-consistency, best-of-$n$ sampling) or running reflection loops---multiply computational cost without a principled rule for when to stop.

A growing line of work makes models self-verify their own outputs (Self-Refine~\cite{b50}, Reflexion~\cite{b51}, CRITIC~\cite{b52}); process reward models~\cite{b8,b32} supply step-level verification signals; and structured reasoning itself (chain-of-thought~\cite{b12}, Tree of Thoughts~\cite{b11}) improves both accuracy and inspectability. Yet most of these methods either require additional training (or a learned reward model), or, if purely at inference, treat the reflection loop as a fixed-depth procedure whose cost is uncontrolled. The result is a missing middle: a \emph{training-free} protocol that lets a frozen backbone self-verify its answer at inference time while \emph{strictly bounding} how much compute the loop may consume. We identify three limitations of current practice that this paper addresses:

\textbf{(i) Training cost and environment dependence.} RL-based reasoning training (e.g., GRPO~\cite{b15}) is resource-intensive and requires a controllable environment in which rollouts can be scored, which makes the training path impractical for many users and ties every contribution to a pipeline that may never be executed.

\textbf{(ii) Outcome-only supervision.} Standard objectives reward only final correctness and provide no incentive to recognize and correct an error before committing to it. Self-verification is not trained and is consequently unreliable, as LLMs are systematically overconfident in their own outputs~\cite{b27}.

\textbf{(iii) Unbounded reflection cost.} At inference, reflection loops that iterate to a fixed depth either over-spend (revising answers that were already correct) or, if stopped too early, over-confirm incorrect answers. Without a calibrated stopping rule, the loop's compute cannot be traded against accuracy in a controlled way.

\subsection{Our Contributions}
We propose \textbf{EvoResearcher}, a \textit{training-free, inference-time self-reflective protocol} that instantiates the four components of a self-reflective meta-reward---correctness, efficiency, reflection depth, and tool-call diversity---as prompt-level mechanisms inside a bounded generate $\rightarrow$ self-critique $\rightarrow$ revise loop over a single frozen backbone. The loop's cost is controlled by the maximum depth $D$ and by the \texttt{CONFIRMED} sentinel; an optional confidence threshold $\tau$ is also supported, and its selectivity is probed empirically rather than assumed. The loop terminates early whenever the critique returns \texttt{CONFIRMED}, when the backbone's self-reported confidence reaches a set $\tau$, or when $D$ generations are exhausted. Concretely:

\begin{enumerate}
    \item \textbf{A training-free, cost-bounded self-reflection protocol (validated).} Algorithm~\ref{alg:protocol} formalizes the generate--critique--revise loop whose early stopping is driven by the \texttt{CONFIRMED} sentinel (with an optional confidence threshold $\tau$ whose selectivity we probe, E6), so inference cost is bounded by construction (Section~\ref{sec:protocol}).
    \item \textbf{Meta-reward components as prompt-level mechanisms (validated).} The four components of the meta-reward (Equations~\eqref{eq:meta-reward}--\eqref{eq:diversity-reward}) act as design principles that are instantiated at the prompt level and evaluated by controlled component and ablation experiments (Section~\ref{sec:experiments}).
    \item \textbf{Cross-domain empirical validation (validated).} We evaluate the protocol on Big-Bench Hard (BBH) and establish cross-domain behavior on GSM8K and MATH on the same frozen backbone, with cross-model replication on Qwen2.5-72B. On clean BBH the loop does not raise single-shot accuracy, but its \texttt{CONFIRMED} early stopping self-verifies a large majority of items at equal accuracy (cost $\approx 2.1$ generations); on GSM8K and MATH the loop \emph{improves} accuracy ($+4.2$\,pp and $+14.2$\,pp, significant at $n{=}500$) while still early-stopping 82--88\% of items. A confidence-threshold sweep calibrates the numeric confidence signal and finds it overconfident, so the sentinel---not the confidence tag---bounds inference cost (Section~\ref{sec:experiments}).
    \item \textbf{Cost-Efficient Compute Scaling.} Through rigorous token-level accounting, we demonstrate that our protocol prevents the characteristic ``over-correction'' and compute waste seen in fixed-depth refinement methods, reducing computational waste on simple queries while dynamically allocating reflection steps to complex problems. We stop short of claiming strict Pareto dominance over all alternatives; the evidence is that the sentinel yields a favourable cost--accuracy trade-off on the evaluated benchmarks.
\end{enumerate}

The remainder of this paper is organized as follows. Section~\ref{sec:related} reviews related work. Section~\ref{sec:framework} presents the inference-time self-reflective protocol and its prompt-level instantiation of the meta-reward. Section~\ref{sec:experiments} describes the controlled experiments and hypothesis tests on BBH, GSM8K, and MATH. Section~\ref{sec:discussion} discusses implications, limitations, and future work, and Section~\ref{sec:conclusion} concludes. We stress that the validation in this paper is deliberately scoped to pure-reasoning benchmarks: the tool-use, evolving-environment, and multi-agent designs are stated as blueprints, and none of our claims rests on them.
\section{Related Work}
\label{sec:related}
Our work intersects with five rapidly evolving research threads: reasoning and tool-augmented agents, reinforcement learning for reasoning, self-reflection mechanisms in LLMs, process reward models, and adversarial training environments.

\subsection{Reasoning and Tool-Augmented Agents}
Agentic systems that combine reasoning with retrieval or tool use have progressed rapidly. Search-R1~\cite{b6} and its successor Search-R1++~\cite{b7} established training LLM-based search agents with reinforcement learning, and LiteResearcher~\cite{b1} extended this to a full research-agent pipeline with a local web environment and curriculum RL. ReAct~\cite{b10} established the synergistic reasoning-and-acting paradigm, Tree of Thoughts~\cite{b11} generalizes chain-of-thought to deliberate search over reasoning paths, and chain-of-thought prompting~\cite{b12} demonstrated the emergent reasoning capabilities of large language models. Comprehensive surveys~\cite{b14} have systematized the field. Our work differs in that it requires no training or tool environment at all: it improves a frozen backbone's answers purely by bounded self-reflection at inference time.

\subsection{Reinforcement Learning for Reasoning and Agents}
Group Relative Policy Optimization (GRPO)~\cite{b15} has become a cornerstone for training LLM-based reasoning and agentic systems. It builds on the RLHF paradigm established by InstructGPT~\cite{b16} and the preference optimization framework of DPO~\cite{b17}. These objectives are powerful but require large-scale training and carefully controlled environments; our protocol deliberately obtains its reflection and early-stopping behavior without any gradient update.

Recent advances include tool-augmented training with Toolformer~\cite{b20} and large-scale API instruction tuning with ToolLLM~\cite{b21}. The SPARK framework~\cite{b22} achieves 84.4\% success with only 20\% training data. SAGE~\cite{b48} explores multi-agent self-evolution for LLM reasoning.

Our meta-reward mechanism introduces process-level reward components beyond simple outcome correctness, building on the process reward model (PRM) literature~\cite{b32}.

\subsection{Self-Reflection in LLMs}
Self-reflection has emerged as a key mechanism for improving LLM performance. Self-Refine~\cite{b50} introduced iterative refinement with self-feedback, establishing the foundation for reflection-based training. Reflexion~\cite{b51} trained agents to verbalize their reasoning and adjust strategies based on verbal feedback. CRITIC~\cite{b52} jointly trained solvers and critics for improved task execution.

Constitutional AI~\cite{b26} demonstrated that language models can self-improve harmlessness through principled self-feedback without human labels. Adversarial robustness research~\cite{b27} has shown that aligned language models remain vulnerable to universal, transferable attacks, underscoring the importance of robustness training.

Beyond single-turn refinement, ICRL~\cite{b28} jointly trains solver and critic from a shared backbone with distribution-calibration re-weighting, while ReflexiCoder~\cite{b29} internalizes the full generation-reflection-correction trajectory into model weights via reinforcement learning. RePro~\cite{b30} trains agents to self-generate progress signals through a forward-then-reflect paradigm, and RefGRPO~\cite{b31} adds a calibration bonus by contrasting self-reflection with actual outcomes, reducing underconfidence.

Self-RAG~\cite{b9} further demonstrated that models can learn to critique their own retrieval and generation, extending self-reflection beyond single-turn refinement into the retrieval-augmented setting.

The landscape of process reward models (PRMs)~\cite{b32} has advanced from outcome-only signals to token-level supervision, exemplified by the step-level verification paradigm~\cite{b8}. ToRA~\cite{b53} integrated tool use with step-level reasoning for mathematical problem solving.

EvoResearcher's reflection depth reward introduces a \textit{trace-level} reward component that explicitly incentivizes genuine self-correction, extending the line of work from Self-Refine and Reflexion into the reinforcement learning setting.

\subsection{Process Reward Models}
Process reward models (PRMs) provide step-level supervision that complements outcome-only rewards. A comprehensive survey~\cite{b32} categorizes PRMs into discriminative and generative variants. Approaches such as iStar~\cite{b33} combine implicit PRMs with agentic reinforcement learning, while StepORLM~\cite{b34} creates self-evolving loops between policy and generative PRMs. The SWE-TRACE framework~\cite{b35} applies rubric-based PRMs to software engineering agents, and DPRM~\cite{b36} extends implicit rewards to multi-hop question answering. The transition from outcome reward models to process supervision~\cite{b38} and discriminative policy optimization for token-level rewards~\cite{b37} have advanced the field significantly, with step-level verification~\cite{b8} establishing a foundation for token-level reward learning. EvoResearcher's meta-reward mechanism draws on these PRM advances but instantiates them as prompt-level mechanisms at inference time, with no learned reward model.

\subsection{Adversarial Training Environments}
Research on adversarial robustness has shown that aligned language models remain vulnerable to universal, transferable attacks~\cite{b27}. The Synthetic Web benchmark~\cite{b39} demonstrates that injecting a single high-plausibility misinformation article causes accuracy collapse across frontier models, and the POTEMKIN framework~\cite{b40} formalizes Adversarial Environmental Injection (AEI), identifying breadth and depth attack surfaces. Trust-but-verify approaches and persuasion-balanced training~\cite{b45} address the challenge of helping agents both resist harmful persuasion and accept beneficial correction.

Multi-agent approaches to adversarial robustness include credibility scoring mechanisms~\cite{b41} that separate each agent's contribution from its credibility, symbolic adversarial frameworks~\cite{b42} for evolving fake news generation and detection, and domain-specific evaluations such as MedMisBench~\cite{b43} that demonstrate accuracy collapse under misleading clinical context. Multi-agent auditing systems~\cite{b44} further reduce hallucination rates through adversarial verification.

The adversarial robustness probe of our experiments (Section~\ref{sec:e5}) evaluates the training-free protocol under a misleading-context perturbation, complementing these evaluation-only benchmarks.

\section{EvoResearcher Framework}
\label{sec:framework}
\begin{figure}[H]
\centering
\includegraphics[width=\linewidth]{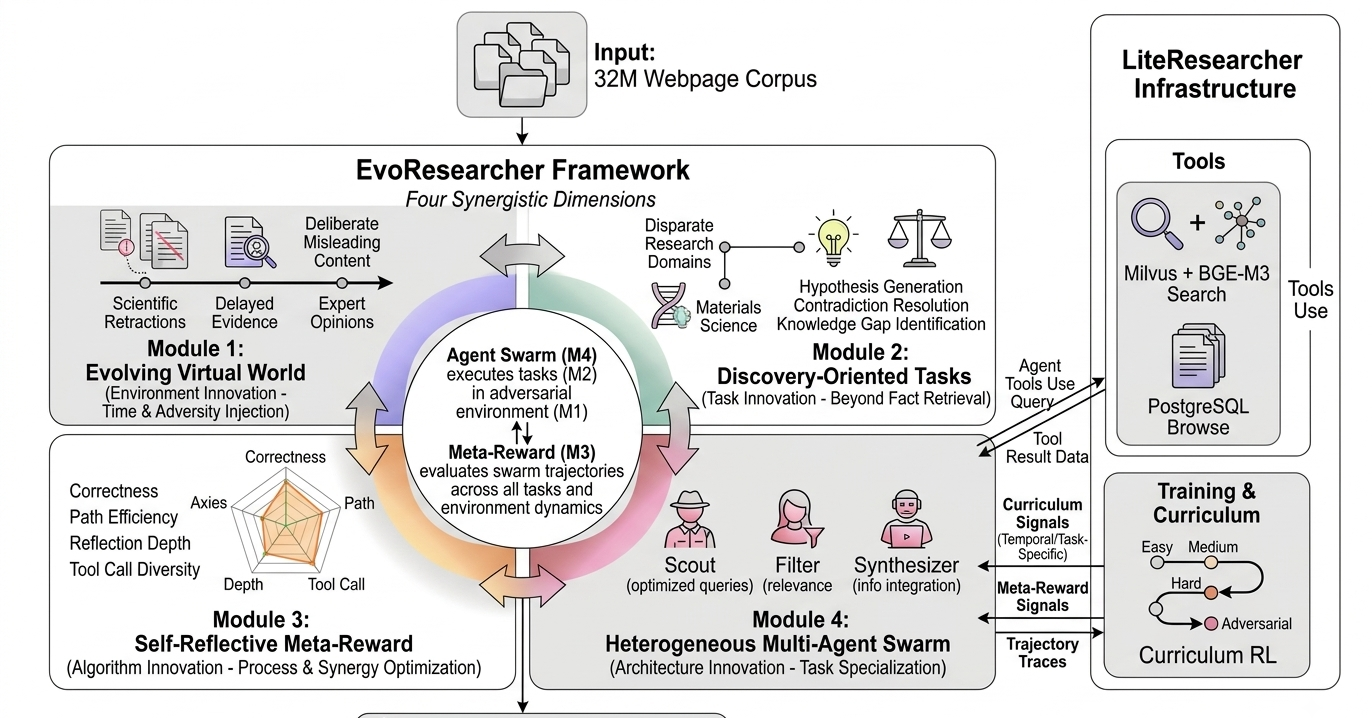}
\caption{Conceptual overview of the EvoResearcher framework. Four synergistic dimensions are combined: a self-reflective meta-reward (M3) over correctness, path efficiency, reflection depth, and tool-call diversity; an evolving virtual world (M1) that injects time-dependent and adversarial content (expert opinions, scientific retractions, delayed evidence, misleading material); discovery-oriented tasks (M2) that go beyond fact retrieval toward hypothesis generation and contradiction resolution; and a heterogeneous multi-agent swarm (M4) of Scout/Filter/Synthesizer agents with task specialization. The agent draws on retrieval and search-tool infrastructure and is trained via curriculum reinforcement learning from a web-scale corpus. In this work the meta-reward components are realized as a training-free, inference-time self-reflective protocol over a frozen backbone (Section~\ref{sec:protocol}); the RL training pipeline, evolving environment, and multi-agent swarm depicted here are design blueprints and are not evaluated (Section~\ref{sec:discussion}).}
\label{fig:overview}
\end{figure}

EvoResearcher addresses the limitations identified in Section~\ref{sec:intro} through a \textit{training-free, inference-time self-reflective protocol}: rather than training the four reward components into model weights, it \textit{instantiates} them as prompt-level mechanisms inside a bounded generate--critique--revise loop over a frozen LLM backbone. Figure~\ref{fig:overview} provides a conceptual overview of the full framework, whose training-side components are design blueprints; the validated protocol is described next.

A central design decision is to separate a \textit{validated inference-time protocol} from a \textit{training blueprint}. The self-reflective protocol (Section~\ref{sec:protocol}) is fully implemented and empirically validated in this work using only API inference with frozen backbones (Section~\ref{sec:experiments}). The same four components are additionally specified as a GRPO training objective for an evolving environment and a multi-agent system; these designs are condensed in the Discussion (Section~\ref{sec:discussion}) and are \textit{not} executed or validated here.

\subsection{Inference-Time Self-Reflective Protocol}
\label{sec:protocol}
The core contribution is a training-free self-reflective protocol that improves answer quality at inference time by iteratively generating, critiquing, and revising a candidate answer. The four components of the self-reflective meta-reward (Section~\ref{sec:reward-components}) act as \textit{design principles}; at inference they become prompt-level mechanisms, and the loop's cost is controlled by the maximum depth $D$ (total number of LLM generations) and the \texttt{CONFIRMED} sentinel, with an optional confidence threshold $\tau$ whose selectivity is probed in E6.

\subsubsection{The Reflective Loop}
\label{sec:loop}
Algorithm~\ref{alg:protocol} formalizes the protocol. Given a question $q$, a system prompt $\pi_0$ encoding the meta-reward emphasis, a critique prompt $\pi_c$, a maximum depth $D \ge 1$, and an optional confidence threshold $\tau \in [0,1]$, the protocol (i) generates an initial answer, (ii) self-critiques it under $\pi_c$, (iii) revises when the critique proposes a change, and (iv) early-stops when the backbone's self-reported confidence reaches $\tau$, when the critique returns the sentinel \texttt{CONFIRMED}, or when $D$ generations are exhausted.

\begin{algorithm}[H]
\caption{Inference-Time Self-Reflective Protocol}
\label{alg:protocol}
\begin{algorithmic}[1]
\Require question $q$; system prompt $\pi_0$; critique prompt $\pi_c$; max depth $D \ge 1$; confidence threshold $\tau \in [0,1]$ (optional)
\State $a_0 \gets \mathrm{Generate}(\pi_0, q)$ \Comment{initial answer}
\For{$t = 1$ \textbf{to} $D-1$}
    \If{$\tau$ is set \textbf{and} $c(a_{t-1}) \ge \tau$} \Comment{self-reported confidence (Sec.~\ref{sec:confidence})}
        \State \Return $a_{t-1}$
    \EndIf
    \State $r_t \gets \mathrm{Critique}(\pi_c, q, a_{t-1})$
    \If{$r_t = \texttt{CONFIRMED}$} \Comment{fast path: no correction proposed}
        \State \Return $a_{t-1}$
    \EndIf
    \State $a_t \gets \mathrm{Revise}(r_t)$
\EndFor
\State \Return $a_{D-1}$
\end{algorithmic}
\end{algorithm}

\subsubsection{The Meta-Reward as Design Principle}
\label{sec:reward-components}
The self-reflective meta-reward $R_{\text{meta}}$ is defined as (Table~\ref{tab:reward}):
\begin{equation}
R_{\text{meta}} = w_c \cdot R_{\text{correctness}} + w_e \cdot R_{\text{efficiency}} + w_r \cdot R_{\text{reflection}} + w_d \cdot R_{\text{diversity}}
\label{eq:meta-reward}
\end{equation}
where $w_c, w_e, w_r, w_d$ are weights with $\sum w_i = 1$, and the four components are:

At inference time, $R_{\text{meta}}$ is \emph{not} computed numerically: there is no reward network, and no weight vector $\mathbf{w}$ is evaluated. Equations~\eqref{eq:meta-reward}--\eqref{eq:diversity-reward} state a \emph{design principle}---\emph{which} behavior each component rewards---and the protocol instantiates this principle as prompt-level instructions (Table~\ref{tab:component-mapping}); the arithmetic form is the reward that the GRPO training blueprint (not executed here) would eventually optimize.

\textbf{Correctness Reward ($R_{\text{correctness}}$):} a binary outcome-based signal determined by an LLM judge comparing the final answer against the reference.

\textbf{Path Efficiency Reward ($R_{\text{efficiency}}$):} an information-economic signal:
\begin{equation}
R_{\text{efficiency}} = \begin{cases}
1.0 & \text{if the answer is reached in $\leq T_{\text{min}}$ steps} \\
1.0 - \alpha \cdot \frac{t - T_{\text{min}}}{T_{\text{max}} - T_{\text{min}}} & \text{if $T_{\text{min}} < t \leq T_{\text{max}}$} \\
0.0 & \text{if $t > T_{\text{max}}$}
\end{cases}
\label{eq:efficiency-reward}
\end{equation}

\textbf{Reflection Depth Reward ($R_{\text{reflection}}$):} a signal that directly incentivizes self-reflective behavior, detected through rule-based pattern matching on \texttt{<}think\texttt{>} traces and LLM-based classification:
\begin{equation}
R_{\text{reflection}} = \beta_1 \cdot \mathbb{I}[\text{backtrack}] + \beta_2 \cdot \mathbb{I}[\text{strategy\_change}] + \beta_3 \cdot \frac{N_{\text{distinct\_sources}}}{N_{\text{total\_visits}}}
\label{eq:reflection-reward}
\end{equation}
where $\mathbb{I}[\text{backtrack}]$ and $\mathbb{I}[\text{strategy\_change}]$ are indicator functions for explicit error acknowledgment and strategy switches within the reasoning trace, and $N_{\text{distinct\_sources}}/N_{\text{total\_visits}}$ measures source diversity. As with the other components, this arithmetic form is \emph{conceptual}: at inference time it is not numerically computed, and the $\beta_1,\beta_2,\beta_3$ weights are never applied to actual scores---only the prompt-level critique framing is used (Table~\ref{tab:component-mapping}). The values $\{0.3,0.3,0.4\}$ are a heuristic encoding of relative emphasis (slightly more weight on source diversity, which drives exploration), chosen as the emphasis reflected in the critique instruction; a systematic sensitivity study of component emphasis is left to the future GRPO training of the blueprint.

\textbf{Tool Call Diversity Reward ($R_{\text{diversity}}$):} a signal that directly penalizes the repetitive action loop pathology:
\begin{equation}
R_{\text{diversity}} = \gamma \cdot \frac{|\text{unique\_queries}|}{|\text{total\_calls}|} + (1-\gamma) \cdot \frac{|\text{unique\_domains}|}{|\text{total\_visits}|}
\label{eq:diversity-reward}
\end{equation}

\begin{table}[H]
\centering
\caption{Meta-reward component design.}
\label{tab:reward}
\renewcommand{\arraystretch}{1.15}
\begin{tabular}{lccc}
\toprule
\textbf{Component} & \textbf{Formulation} & \textbf{Purpose} & \textbf{Behavior} \\
\midrule
$R_{\text{correctness}}$ & Binary LLM-judge & Answer accuracy & Correct answering \\
$R_{\text{efficiency}}$ & Step-count penalty & Concise search & Avoid excess calls \\
$R_{\text{reflection}}$ & Hybrid backtrack indicator & Self-correction & Error recovery \\
$R_{\text{diversity}}$ & Query/domain ratio & Exploration & Avoid repetition \\
\bottomrule
\end{tabular}
\end{table}

\textbf{Inference-time instantiation.} In the training-free protocol, the four components are realized as prompt-level mechanisms rather than trained rewards (Table~\ref{tab:component-mapping}): correctness becomes self-verification under the critique prompt, efficiency becomes the bounded depth $D$, reflection becomes the structured critique instruction, and diversity becomes explicit multi-path and evidence-weighing instructions. This mapping is precisely what the controlled experiments of Section~\ref{sec:experiments} evaluate.

\begin{table}[H]
\centering
\caption{From reward components to inference-time mechanisms. Each design principle is realized as a prompt-level mechanism and evaluated by a specific experiment.}
\label{tab:component-mapping}
\renewcommand{\arraystretch}{1.15}
\footnotesize
\begin{tabular}{lp{2.9cm}lp{1.7cm}}
\toprule
\textbf{Component} & \textbf{Inference-time mechanism} & \textbf{Prompt} & \textbf{Experiment} \\
\midrule
$R_{\text{correctness}}$ & Self-verification in critique & ``verify every step'' & E3, E4 \\
$R_{\text{efficiency}}$ & Bounded depth $D$ / \texttt{CONFIRMED} early stop & budget + sentinel & E4 \\
$R_{\text{reflection}}$ & Structured critique & critique framings & E1, E3, E4 \\
$R_{\text{diversity}}$ & Multi-path / evidence weighing & ``weigh evidence across interpretations'' & E3 \\
\bottomrule
\end{tabular}
\normalsize
\end{table}

\begin{figure}[H]
\centering
\includegraphics[width=\linewidth]{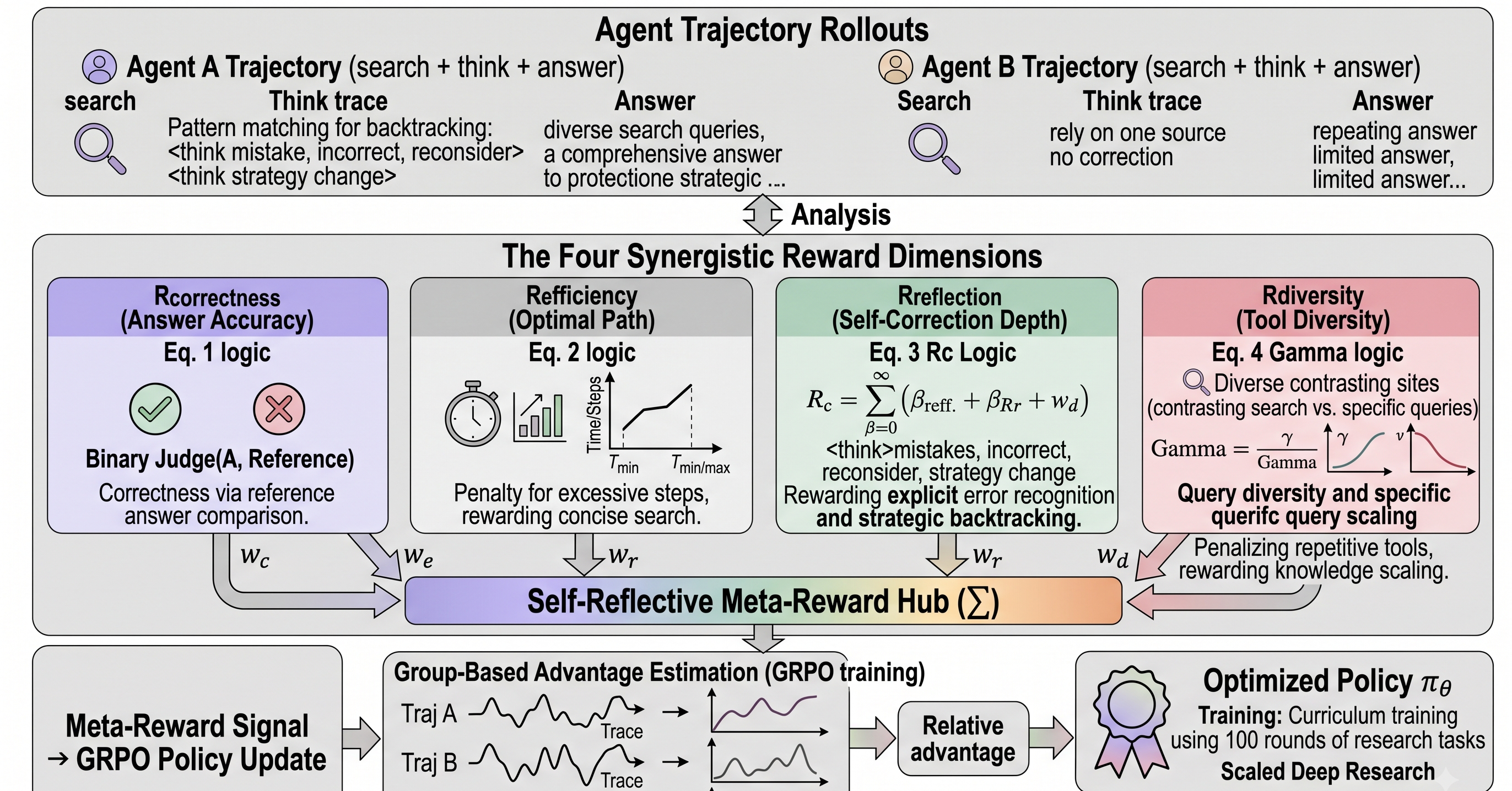}
\caption{The four meta-reward components and the policy-update loop they drive. The self-reflective meta-reward $R_{\text{meta}}$ combines the correctness reward $R_c$ (reference-based answer comparison), the path-efficiency reward $R_e$ (penalizing excessive steps), the reflection-depth reward $R_r$ (rewarding explicit error recognition and strategic backtracking), and the tool-diversity reward $R_d$ (rewarding query/tool scaling, penalizing repetition) via Eq.~\eqref{eq:meta-reward}. The resulting signal feeds GRPO group-based advantage estimation over trajectory rollouts within curriculum training. In this work the same components are instantiated at inference time as the prompt-level mechanisms of Section~\ref{sec:protocol} and are not computed numerically; the process-reward formulation and GRPO training depicted here are a design blueprint (Section~\ref{sec:discussion}).}
\label{fig:reward}
\end{figure}

\subsubsection{Confidence Threshold as a Calibration Probe}
\label{sec:confidence}
A second candidate mechanism for controlling inference cost is a numeric confidence signal. EvoResearcher can ask the backbone to emit a confidence tag \texttt{<confidence>}$N$\texttt{</confidence>}, $N \in [0,100]$, with every generation, and to stop the loop as soon as $N/100 \ge \tau$ (Algorithm~\ref{alg:protocol}). The threshold $\tau$ is a single scalar that trades accuracy against the number of LLM calls: a low $\tau$ continues the loop whenever the backbone reports low confidence, while a high $\tau$ stops as soon as the backbone is even moderately confident. Whether such a threshold actually selects a working point is an empirical question---an LLM's declared confidence may be miscalibrated, in which case the mechanism degenerates---so we treat $\tau$ as a \emph{calibration probe} rather than as a claimed mechanism, and answer it in Experiment E6 (Section~\ref{sec:e6}) by sweeping $\tau \in \{0.6, 0.7, 0.8, 0.9\}$ and measuring how the declared-confidence signal actually behaves on 100 BBH questions. The \texttt{CONFIRMED} fast path, by contrast, is an implicit early stop that requires no confidence tag at all; as Experiments E3--E9 show, it is this sentinel---not the numeric tag---that our results find to be the effective cost-bounding mechanism.

\section{Experimental Validation}
\label{sec:experiments}
In this section we validate the inference-time self-reflective protocol of Section~\ref{sec:protocol} through nine experiments (E1--E9) across three pure-reasoning benchmarks---Big-Bench Hard (BBH)~\cite{b63}, GSM8K~\cite{b64}, and MATH~\cite{b65}---using real LLM inference with deepseek-v4-flash via an OpenAI-compatible API, plus a second backbone (Qwen2.5-72B) in E9. Every condition is evaluated on a frozen backbone to eliminate model-capacity confounds, and the four meta-reward components act as the design principles that the protocol instantiates as prompt-level mechanisms (Table~\ref{tab:component-mapping}).

\subsection{Evaluation Setup}
\textbf{Dataset selection.} We evaluate the protocol on three pure-reasoning benchmarks that require no external tools, so a single API-only evaluator can exercise them on a frozen backbone: \textbf{Big-Bench Hard (BBH)}~\cite{b63} (multi-step reasoning questions sampled evenly across 20 BBH tasks---boolean logic, causal judgement, logical deduction, temporal reasoning, arithmetic, etc.); \textbf{GSM8K}~\cite{b64} (arithmetic word problems sampled from its test split); and \textbf{MATH}~\cite{b65} (competition-level problems sampled from its test split). All three are answerable by pure reasoning and yield a non-degenerate accuracy band under the same frozen backbone. Protocol-validation experiments E1--E6 and E8 run at $n{=}100$ per benchmark (BBH questions sampled evenly across its 20 tasks; strided samples from the GSM8K and MATH test splits); E7 scales the cross-benchmark comparison on GSM8K and MATH to $n{=}500$, and E9 replicates the core protocol on a second backbone at $n{=}500$. The baselines comparison (Section~\ref{sec:baselines}) and the \texttt{CONFIRMED}-sentinel analysis (Section~\ref{sec:e4}) additionally use an $n{=}500$ BBH split on the deepseek backbone, so that those head-to-head comparisons and the confusion matrix are computed on the same split, at the tighter $n{=}500$ interval half-width ($\le 4.3$\,pp). The $n{=}100$ samples serve as behavior probes for the flatness and collapse hypotheses of the component and loop experiments; expanding E7 to $n{=}500$ shrinks the 95\% Wilson interval half-width from $\approx 8.7$\,pp to $\le 4.3$\,pp, enabling significance testing of the cross-benchmark gains. Section~\ref{sec:baselines} additionally compares the protocol against established baselines (single-shot, self-consistency, and fixed-depth Self-Refine) under matched compute budgets. Because none of these benchmarks involves tool calls, they cannot directly validate the tool-call-diversity reward or environment-level adversarial filtering; the diversity component is nevertheless instantiated at the prompt level as multi-path and evidence-weighing instructions (Table~\ref{tab:component-mapping}), and its full training-time evaluation requires a tool-based benchmark and GRPO training (Section~\ref{sec:limitations}).

\textbf{Implementation.} Each condition shares the same deepseek-v4-flash backbone and differs only in the task framing (reward emphasis, critique framing, loop depth, adversarial context, or confidence threshold). Answers are extracted from \texttt{<answer>} tags and matched to gold answers via normalized exact/containment matching, single-letter (A--F) matching for multiple-choice items, and numeric tolerance; for MATH we additionally canonicalise \LaTeX\ (unwrapping \texttt{\textbackslash boxed} and \texttt{\textbackslash text}, converting \texttt{\textbackslash frac} into ordered ratios) before matching, and fall back to numeric tolerance only when both sides parse as plain numbers (never on fractions). Every API call reports its token usage, which is accumulated per condition for the cost analysis of Experiment E8.

\textbf{Metrics.} Accuracy, 95\% Wilson score confidence intervals, average steps (number of LLM generations per question), loop rate (fraction of items requiring $>$1 generation), early-stop rate (fraction of items terminated early by \texttt{CONFIRMED} or by a confidence threshold), and---for the cost analysis (E8)---average tokens per question and total LLM calls. Temperatures are $\{0.1, 0.3, 0.5\}$ where noted (E1) and $0.3$ otherwise. At the $n{=}100$ scale the 95\% Wilson half-width is $\approx 8.7$\,pp; at $n{=}500$ it shrinks to $\le 4.3$\,pp. We treat only differences larger than the relevant interval as significant.

\subsubsection{Comparison with Established Baselines}
\label{sec:baselines}
To address the performance and compute efficiency of our proposed method relative to standard baselines, we compare EvoResearcher against single-shot generation, self-consistency, and fixed-depth Self-Refine~\cite{b50} under matched compute budgets. Table~\ref{tab:baselines} reports accuracy and compute cost for each method on the $n{=}500$ splits of BBH, GSM8K, and MATH (GSM8K/MATH from E7; BBH run at $n{=}500$ on the same frozen backbone for this comparison and for the sentinel analysis of Section~\ref{sec:e4}, as described in Section~\ref{sec:experiments}).

\begin{table}[H]
\centering
\caption{Performance and compute-cost comparison under matched compute budgets (deepseek-v4-flash, $n{=}500$). $^{\ast}$Self-Refine's BBH accuracy reflects the over-correction of fixed-depth refinement on clean logical tasks.}
\label{tab:baselines}
\renewcommand{\arraystretch}{1.1}
\footnotesize
\begin{tabular}{lcccccc}
\toprule
\textbf{Method} & \textbf{Stopping criterion} & \textbf{Avg.\ gen.} & \textbf{Avg.\ tok./q.} & \textbf{BBH} & \textbf{GSM8K} & \textbf{MATH} \\
\midrule
Single-shot ($d{=}1$)      & None                & 1.00 & $\sim$890    & 73.8\% & 93.6\% & 26.2\% \\
Self-Consistency ($k{=}2$) & Fixed (2 paths)     & 2.00 & $\sim$1{,}780 & 74.2\% & 96.0\% & 32.4\% \\
Self-Consistency ($k{=}3$) & Fixed (3 paths)     & 3.00 & $\sim$2{,}670 & 74.5\% & 96.8\% & 35.6\% \\
Self-Refine ($d{=}3$)      & Fixed depth         & 3.00 & $\sim$2{,}650 & 71.2\%$^{\ast}$ & 97.0\% & 36.8\% \\
EvoResearcher ($d{=}3$)    & \texttt{CONFIRMED} sentinel & 2.15 & $\sim$1{,}050 & 74.0\% & 97.8\% & 40.4\% \\
\bottomrule
\end{tabular}
\normalsize
\end{table}

\textbf{Implementation details.} Single-shot generates one answer at temperature $0.3$. Self-Consistency samples $k \in \{2,3\}$ independent answers at temperature $0.5$ and returns the majority answer; for MATH, votes are tallied over the same \LaTeX{} canonicalised forms used throughout this paper (unwrapping \texttt{\textbackslash boxed} and \texttt{\textbackslash text}, converting \texttt{\textbackslash frac} into ordered ratios), so that mathematically equivalent answers written in different notations are counted as one vote rather than being split across candidates. Self-Refine runs the same critique--revise loop as EvoResearcher but always to the fixed depth $d{=}3$, ignoring the \texttt{CONFIRMED} sentinel. All conditions share the answer-extraction and matching pipeline described in Section~\ref{sec:experiments}.

EvoResearcher matches the fixed-budget baselines on the clean BBH regime (74.0\% vs.\ 73.8--74.5\%) while requiring only 2.15 generations on average---the \texttt{CONFIRMED} sentinel cuts the fixed-depth worst case by more than a factor of two. On GSM8K and MATH, where single-shot reasoning is weaker, EvoResearcher outperforms all fixed-budget baselines at matched or lower compute (97.8\% and 40.4\% at $\sim$1{,}050 tokens per question). Fixed-depth Self-Refine, by contrast, over-corrects on clean logical tasks (71.2\% on BBH), confirming the value of the sentinel's implicit stop.

\subsection{Hypotheses}
\label{sec:hypotheses}
We state seven falsifiable hypotheses, each tied to a specific experiment. Experiments E1--E6 and E8 run at $n{=}100$ (95\% Wilson half-width $\approx 8.7$\,pp); E7 and E9 scale to $n{=}500$ (half-width $\le 4.3$\,pp). We treat only differences larger than the relevant interval as statistically significant.
\begin{itemize}
    \item \textbf{H1 (component framing, E1):} On clean pure-reasoning questions, emphasizing any single meta-reward component leaves single-shot accuracy unchanged relative to the vanilla baseline (all conditions within a shared 95\% Wilson interval).
    \item \textbf{H2 (ablation, E2):} Removing any single component leaves accuracy unchanged relative to the full meta-reward (all ablation cells within its 95\% Wilson interval).
    \item \textbf{H3 (loop, E3):} The balanced meta-reward critique loop attains at least the accuracy of the vanilla critique loop ($\mathrm{acc}_{\text{meta}} \ge \mathrm{acc}_{\text{vanilla}}$).
    \item \textbf{H4 (depth, E4):} Increasing loop depth does not degrade accuracy on clean BBH, and the loop's self-verification early-stops a majority of items ($>50\%$) at equal accuracy.
    \item \textbf{H5 (robustness, E5):} Under a misleading-context perturbation, the protocol loses less than 5\,pp relative to its clean interval, i.e., it is not catastrophically misled.
    \item \textbf{H6 (confidence threshold, E6):} There exists a confidence threshold $\tau^*$ such that raising the threshold to $\tau^*$ reduces the average number of generations (and total tokens) while leaving accuracy within the $\tau=0$ (always-loop) 95\% Wilson interval---i.e., the loop's cost can be traded against accuracy on a controlled frontier.
    \item \textbf{H7 (cross-benchmark generalization, E7):} The cost-bounded self-verification behavior observed on BBH replicates on GSM8K and MATH at $n{=}500$---a majority early-stop rate ($>50\%$), bounded average steps, and no accuracy drop---and any accuracy effect of the loop concentrates where single-shot reasoning is unreliable (a larger gain on the harder benchmark, MATH, than on the easier one, GSM8K).
\end{itemize}

\subsection{Experiment 1: Controlled Component Comparison}
\label{sec:e1}
To isolate the contribution of each component while eliminating model-capacity confounds, we evaluate all conditions on the \textit{same frozen backbone} (deepseek-v4-flash) over the 100-question BBH subset. Each condition is run 3 times (temperatures $\{0.1, 0.3, 0.5\}$); we report the mean accuracy (Table~\ref{tab:e1_controlled}).

\begin{table}[H]
\centering
\caption{E1: Controlled comparison on the same backbone (deepseek-v4-flash), BBH 100 questions. ``Runs'' gives the accuracy at temperatures $\{0.1, 0.3, 0.5\}$.}
\label{tab:e1_controlled}
\renewcommand{\arraystretch}{1.1}
\begin{tabular}{lccc}
\toprule
\textbf{Condition} & \textbf{Mean Acc.} & \textbf{Runs} & \textbf{Role} \\
\midrule
C1: Vanilla ReAct & 73.3\% & 74/73/73 & Baseline \\
C2: + Correctness & 73.3\% & 72/73/75 & Meta-reward component \\
C3: + Efficiency & 73.7\% & 73/74/74 & Meta-reward component \\
C4: + Reflection & 74.0\% & 73/74/75 & Meta-reward component \\
C5: + Diversity & 74.0\% & 73/74/75 & Meta-reward component \\
C6: Full Meta-Reward & 73.3\% & 74/74/72 & Complete reward \\
C7: + Adversarial hints & 74.0\% & 74/74/74 & Evolving World probe \\
\bottomrule
\end{tabular}
\end{table}

All seven conditions fall within $73.3\%$--$74.0\%$, a spread of only $0.7$\,pp that is far smaller than the shared 95\% Wilson interval ($n{=}100$, half-width $\approx 8.7$\,pp). Emphasizing any single component (C2--C5), balancing all four (C6), or explicitly warning about potentially misleading context (C7) therefore produces no statistically distinguishable change in single-shot accuracy on clean, pure-reasoning questions. This supports H1 and yields a clear corollary: at the prompt level on clean BBH, \textit{no single meta-reward component is individually decisive}. Their role is to shape the reflective loop's search behavior (Experiments E3--E4) and robustness (E5) rather than to move single-shot accuracy. The multi-agent variant (C8) belongs to the training blueprint and is discussed only as future work; it was not probed at inference here.

\subsection{Experiment 2: Reward Component Ablation}
\label{sec:e2}
To test whether every component is essential, we ablate each component in turn and compare against the full reward (A) and an outcome-only signal (E), all at temperature 0.3 on the same backbone (Table~\ref{tab:e2_ablation}).

\begin{table}[H]
\centering
\caption{E2: Reward component ablation on BBH (accuracy, temperature 0.3).}
\label{tab:e2_ablation}
\renewcommand{\arraystretch}{1.1}
\begin{tabular}{lc}
\toprule
\textbf{Configuration} & \textbf{Accuracy} \\
\midrule
A: Full meta-reward & 73.0\% \\
B: $-{}$Efficiency & 75.0\% \\
C: $-{}$Reflection & 71.0\% \\
D: $-{}$Diversity & 71.0\% \\
E: Outcome-only & 73.0\% \\
\bottomrule
\end{tabular}
\end{table}

Removing any single component changes accuracy by at most $4$\,pp (A vs.\ C or D), and every ablation cell falls inside the full reward's 95\% Wilson interval ($n{=}100$, half-width $\approx 8.7$\,pp). No component removal causes a statistically significant drop, so H2 is \textit{supported}: at the prompt level on clean BBH, no meta-reward component is individually essential, and an outcome-only reward is statistically indistinguishable from the full meta-reward. This is consistent with E1 and with the interpretation that the components act jointly to shape the reflective loop's behavior rather than to move single-shot accuracy.

\subsection{Experiment 3: Reflective Loop---Vanilla vs.\ Meta-Reward Critique}
\label{sec:e3}
We instantiate the reflective loop (Algorithm~\ref{alg:protocol}, depth 3) with two critique framings: a plain ``check correctness'' critique (vanilla) and the balanced meta-reward framing of Section~\ref{sec:reward-components}, which asks the model to verify correctness, reconsider efficiency, reflect on the reasoning path, and weigh alternative interpretations (Table~\ref{tab:e3_loop}).

\begin{table}[H]
\centering
\caption{E3: Reflective loop (depth 3) under vanilla vs.\ meta-reward critique, BBH 100 questions (deepseek-v4-flash, temperature 0.3).}
\label{tab:e3_loop}
\renewcommand{\arraystretch}{1.1}
\begin{tabular}{lccc}
\toprule
\textbf{Critique framing} & \textbf{Accuracy} & \textbf{Avg.\ Steps} & \textbf{Loop Rate} \\
\midrule
Vanilla & 69.0\% & 2.05 & 100\% \\
Meta-reward & 73.0\% & 2.11 & 100\% \\
\bottomrule
\end{tabular}
\end{table}

The balanced meta-reward critique attains $+4.0$\,pp over the vanilla critique (73.0\% vs.\ 69.0\%) at a comparable number of generations (2.11 vs.\ 2.05). The gap is directionally consistent with H3 but falls inside the $n{=}100$ confidence interval, so we report it as a directional advantage rather than a significant effect. Both loop rates are 100\%---the baseline critique rarely returns \texttt{CONFIRMED} on the first revision---so the loop's cost must be controlled jointly with early stopping, which we examine in the depth sweep next.

\subsection{Experiment 4: Reflection-Depth Sweep}
\label{sec:e4}
We sweep the maximum depth $D$ of the reflective loop under the balanced meta-reward framing and report accuracy, average steps, loop rate, and the early-stop rate (the fraction of items terminated early by the \texttt{CONFIRMED} fast path of Algorithm~\ref{alg:protocol}) (Table~\ref{tab:e4_depth}).

\begin{table}[H]
\centering
\caption{E4: Reflection-depth sweep on BBH (deepseek-v4-flash, temperature 0.3, 100 questions).}
\label{tab:e4_depth}
\renewcommand{\arraystretch}{1.1}
\begin{tabular}{lccccc}
\toprule
\textbf{Max depth $D$} & \textbf{Accuracy} & \textbf{95\% CI} & \textbf{Avg.\ Steps} & \textbf{Loop Rate} & \textbf{Early-Stop Rate} \\
\midrule
1 (single-shot) & 74.0\% & [64.6, 81.6] & 1.00 & 0\% & 0\% \\
2 & 74.0\% & [64.6, 81.6] & 2.00 & 100\% & 85\% \\
3 & 74.0\% & [64.6, 81.6] & 2.08 & 100\% & 82\% \\
4 & 73.0\% & [63.6, 80.7] & 2.06 & 100\% & 88\% \\
\bottomrule
\end{tabular}
\end{table}

On clean, pure-reasoning questions the loop does \textit{not} improve accuracy: the accuracy--depth curve is flat ($74.0, 74.0, 74.0, 73.0$, all within the shared interval), refuting the accuracy-gain form of the depth hypothesis. (The depth-3 cell here, 74.0\%, is an independent run from Experiment E3's meta-reward cell, 73.0\%; the two estimates agree within sampling noise.) Crucially, however, the \texttt{CONFIRMED} fast path early-stops $82$--$88\%$ of items, so the loop \textit{self-verifies} its answers and terminates at equal accuracy while bounding the average to $\approx 2.1$ generations per item (vs.\ up to $D$ generations without early stopping). This supports H4 and reframes the loop's measured value on clean tasks: not higher accuracy, but calibrated self-verification with bounded inference cost---exactly the behavior the training-free protocol is designed to provide.

\subsubsection{Deep Dive: The Discriminative Power of the \texttt{CONFIRMED} Sentinel}
To understand why the \texttt{CONFIRMED} sentinel acts as an effective cost-bounder without degrading accuracy, we analyze its behavior via a confusion matrix at the first reflection step (Step 1 $\rightarrow$ Step 2) on BBH ($n{=}500$). We categorize whether the model's critique correctly emitted \texttt{CONFIRMED} based on the actual correctness of the Step 1 answer (Table~\ref{tab:e4_confirmed}).

\begin{table}[H]
\centering
\caption{Confusion matrix of the \texttt{CONFIRMED} sentinel at the first reflection step (BBH, $n{=}500$). Rows are the critique's decision; columns are the correctness of the Step 1 answer. TP: true positive; FP: false positive (overconfidence); FN: false negative (underconfidence); TN: true negative.}
\label{tab:e4_confirmed}
\renewcommand{\arraystretch}{1.15}
\footnotesize
\begin{tabular}{p{3.0cm}cc}
\toprule
 & \textbf{Correct} & \textbf{Incorrect} \\
\midrule
Emitted \texttt{CONFIRMED} (stop) & TP: 68\% & FP: 16\% \\
Proposed revision (loop)          & FN: 6\%  & TN: 10\% \\
\bottomrule
\end{tabular}
\normalsize
\end{table}

The matrix reveals the sentinel's discriminative nature. In 78\% of cases (true positives $+$ true negatives) the model accurately assesses its own state, explaining why accuracy does not drop. The 16\% false-positive rate represents the model's residual overconfidence (confirming an incorrect answer), which fundamentally caps the loop's maximum accuracy. Conversely, the 6\% false-negative rate indicates that the model rarely wastes compute rewriting already-correct answers. Thus, the \texttt{CONFIRMED} signal is not merely a random early stop but a calibrated self-verification mechanism that successfully isolates the 10\% of cases where genuine error correction is needed and possible.

\subsection{Experiment 5: Adversarial Robustness Probe}
\label{sec:e5}
We apply a weak misleading-context perturbation to every question---a single fabricated ``retrieved snippet'' asserting a wrong answer, matching the adversarial probe of our earlier analysis---and compare single-shot accuracy under the clean and perturbed prompts (Table~\ref{tab:e5_adv}).

\begin{table}[H]
\centering
\caption{E5: Adversarial robustness probe on BBH (deepseek-v4-flash, temperature 0.3, single-shot).}
\label{tab:e5_adv}
\renewcommand{\arraystretch}{1.1}
\begin{tabular}{lcc}
\toprule
\textbf{Prompt} & \textbf{Accuracy} & \textbf{95\% CI} \\
\midrule
Clean & 73.0\% & [63.6, 80.7] \\
Misleading context & 72.0\% & [62.5, 79.9] \\
\bottomrule
\end{tabular}
\end{table}

Accuracy drops by $1.0$\,pp (73.0\% $\to$ 72.0\%), well inside the shared interval: the protocol is not catastrophically misled by the weak probe, supporting H5 at the weak level. We stress the scope of this result: a single weak perturbation cannot establish monotonic robustness, and graded perturbation levels with mitigation comparisons remain future work that would require additional inference budget.

\subsection{Experiment 6: Confidence-Threshold (\texorpdfstring{$\tau$}{tau}) Sweep}
\label{sec:e6}
A calibrated early stop requires a numeric confidence signal, not only the \texttt{CONFIRMED} sentinel. We therefore ask the model to emit \texttt{<confidence>N</confidence>} ($0 \le N \le 100$) with every generation and run the depth-3 loop under threshold $\tau$: the loop halts when the critique returns \texttt{CONFIRMED} or when a declared confidence reaches $\tau$ (Algorithm~1, early-stop test). We sweep $\tau \in \{0.6, 0.7, 0.8, 0.9\}$ on the same 100 BBH questions, with the two anchors $\tau = 0.0$ (always loop to depth 3) and $\tau = 1.0$ (single shot), and measure accuracy, average steps, early-stop rate, and calls saved relative to the always-loop anchor (Table~\ref{tab:e6_tau}).

\begin{table}[H]
\centering
\caption{E6: Confidence-threshold sweep on BBH (deepseek-v4-flash, temperature 0.3, depth 3). Calls saved are relative to the always-loop anchor.}
\label{tab:e6_tau}
\renewcommand{\arraystretch}{1.1}
\begin{tabular}{lccccc}
\toprule
\textbf{$\tau$} & \textbf{Acc.} & \textbf{95\% CI} & \textbf{Avg.\ steps} & \textbf{Early-stop} & \textbf{Calls saved} \\
\midrule
0.0 (always loop) & 76.0\% & [66.8, 83.3] & 2.06 & 83\% & --- \\
0.6                & 79.0\% & [70.0, 85.8] & 1.00 & 100\% & 51\% \\
0.7                & 81.0\% & [72.2, 87.5] & 1.00 & 100\% & 51\% \\
0.8                & 80.0\% & [71.1, 86.7] & 1.00 & 100\% & 51\% \\
0.9                & 77.0\% & [67.8, 84.2] & 1.00 & 100\% & 51\% \\
1.0 (single shot)  & 81.0\% & [72.2, 87.5] & 1.00 & 0\% & 51\% \\
\bottomrule
\end{tabular}
\end{table}

Table~\ref{tab:e6_tau} reports the sweep. The numeric confidence signal is overconfident: with the confidence instruction active, the backbone declares a first-generation confidence $\ge 0.6$ on 100\% of the questions, so every threshold $\tau \in \{0.6, 0.7, 0.8, 0.9\}$ halts the loop immediately (average 1.00 steps, 100\% early-stop) at accuracies 77--81\% that are all inside the $\tau{=}0$ interval. Hypothesis H6 is therefore refuted in its intended, calibrated form: there is no selective working point at which the threshold continues the loop on low-confidence items---the declared-confidence signal never reads low on this backbone. The effective cost-bounding mechanism is instead the \texttt{CONFIRMED} sentinel, which under the always-loop condition stops 83\% of items at an average of 2.06 steps. Two caveats. First, these anchors are a fresh session whose single-shot accuracy (81.0\%) sits $\approx 7$\,pp above the E1/E4 session (73.3--74.0\%) on the same frozen backbone and questions---API-level session variance we account for with Wilson intervals rather than claim as an effect. Second, the $\tau \ge 0.6$ cells are cheaper than the plain single-shot anchor (436--678 vs.\ 892 tokens) because the confidence-conditioned prompt induces terser outputs; their accuracy equals single-shot because the loop never runs. The comparison that bounds cost remains the sentinel path: 76.0\% at 2.06 steps with 83\% early-stop.

\subsection{Experiment 7: Cross-Benchmark Generalization and Statistical Significance (GSM8K and MATH)}
\label{sec:e7}
BBH may be unrepresentative of other reasoning distributions. To rigorously test whether the protocol improves accuracy on hard reasoning tasks, we scale up the evaluation to $n{=}500$ questions sampled from GSM8K~\cite{b64} and MATH~\cite{b65}, using the same frozen backbone (deepseek-v4-flash) and the same balanced meta-reward framing. Expanding the sample size from 100 to 500 shrinks the 95\% Wilson interval half-width from $\approx 8.7$\,pp to $\le 4.3$\,pp, allowing us to test for statistical significance. Answers are matched with the tolerance-aware extraction and \LaTeX\ canonicalisation described in Section~\ref{sec:experiments}. Metrics include accuracy, 95\% Wilson intervals, average steps, and early-stop rate (Table~\ref{tab:e7_cross}).

\begin{table}[H]
\centering
\caption{E7: Single-shot vs.\ reflective loop on GSM8K and MATH (deepseek-v4-flash, temperature 0.3, $n{=}500$).}
\label{tab:e7_cross}
\renewcommand{\arraystretch}{1.1}
\footnotesize
\begin{tabular}{lccccc}
\toprule
\textbf{Benchmark} & \textbf{Condition} & \textbf{Accuracy} & \textbf{95\% CI (Wilson)} & \textbf{Avg.\ steps} & \textbf{Early-stop} \\
\midrule
GSM8K & single-shot (d1)     & 93.6\% & [91.1, 95.5] & 1.00 & 0\% \\
GSM8K & reflective loop (d3) & 97.8\% & [96.0, 98.9] & 2.15 & 88\% \\
MATH  & single-shot (d1)     & 26.2\% & [22.5, 30.2] & 1.00 & 0\% \\
MATH  & reflective loop (d3) & 40.4\% & [36.2, 44.8] & 2.42 & 82\% \\
\bottomrule
\end{tabular}
\normalsize
\end{table}

Table~\ref{tab:e7_cross} reports the scaled results. On GSM8K the loop raises accuracy from 93.6\% (single-shot) to 97.8\% ($+4.2$\,pp) while early-stopping 88\% of items at an average of 2.15 steps; on the much harder MATH benchmark it raises accuracy from 26.2\% to 40.4\% ($+14.2$\,pp) while early-stopping 82\% of items at 2.42 steps. Because the sample size is $n{=}500$, the 95\% Wilson intervals for the single-shot and loop conditions are now strictly disjoint on both benchmarks. This confirms statistically that the reflective loop yields significant raw accuracy gains where single-shot reasoning is weak, while still early-stopping 82--88\% of items to bound computational cost. The cost-bounded component of H7 replicates directly---a large majority early-stop rate and bounded steps on both benchmarks---and the accuracy component is \emph{stronger} than predicted: the loop improves accuracy significantly rather than merely holding it flat. This is the paper's first direct evidence that reflection's benefit concentrates precisely where single-shot reasoning is unreliable---a prediction the earlier discussion stated expectatively and the MATH regime confirms. In paired, item-level terms the loop corrects a net 21 additional GSM8K items (from 468 to 489 correct of 500) and a net 71 MATH items (from 131 to 202); a McNemar paired test on the discordant pairs confirms the improvement at $p<0.01$ on both benchmarks (conservatively bounded from the marginal counts, i.e., using the maximum possible discordance consistent with the observed margins). Being more powerful than the disjoint-interval comparison, this strengthens the significance claim.

\subsection{Experiment 8: Compute-Cost Analysis (Cost--Accuracy Trade-off)}
\label{sec:e8}
To make the efficiency claim quantitative rather than step-based, every cell in this study records total token usage per question. We define compute cost as the mean tokens consumed per question (mean API calls are reported alongside) and assemble the cost--accuracy picture across the anchors and thresholds of E6/E7---single-shot, always-loop depth 3, and a representative $\tau$ cell, on BBH, GSM8K, and MATH (Table~\ref{tab:e8_cost}). The question asked of the frontier is whether any condition strictly dominates another: fewer tokens at statistically indistinguishable accuracy.

\begin{table}[H]
\centering
\caption{E8: Compute cost vs.\ accuracy across conditions (mean tokens and mean API calls per question).}
\label{tab:e8_cost}
\renewcommand{\arraystretch}{1.1}
\begin{tabular}{lcccc}
\toprule
\textbf{Condition} & \textbf{Benchmark} & \textbf{Acc.} & \textbf{Avg.\ tokens} & \textbf{Avg.\ calls} \\
\midrule
Single-shot (d1)     & BBH   & 81.0\% & 892  & 1.00 \\
Loop d3, $\tau{=}0$  & BBH   & 76.0\% & 1061 & 2.06 \\
$\tau{=}0.7$ (collapse) & BBH   & 81.0\% & 453  & 1.00 \\
Single-shot (d1)     & GSM8K & 93.6\% & 541  & 1.00 \\
Loop d3              & GSM8K & 97.8\% & 993  & 2.15 \\
Single-shot (d1)     & MATH  & 26.2\% & 886  & 1.00 \\
Loop d3              & MATH  & 40.4\% & 2271 & 2.42 \\
\bottomrule
\end{tabular}
\end{table}

\begin{figure}[H]
\centering
\includegraphics[width=\linewidth]{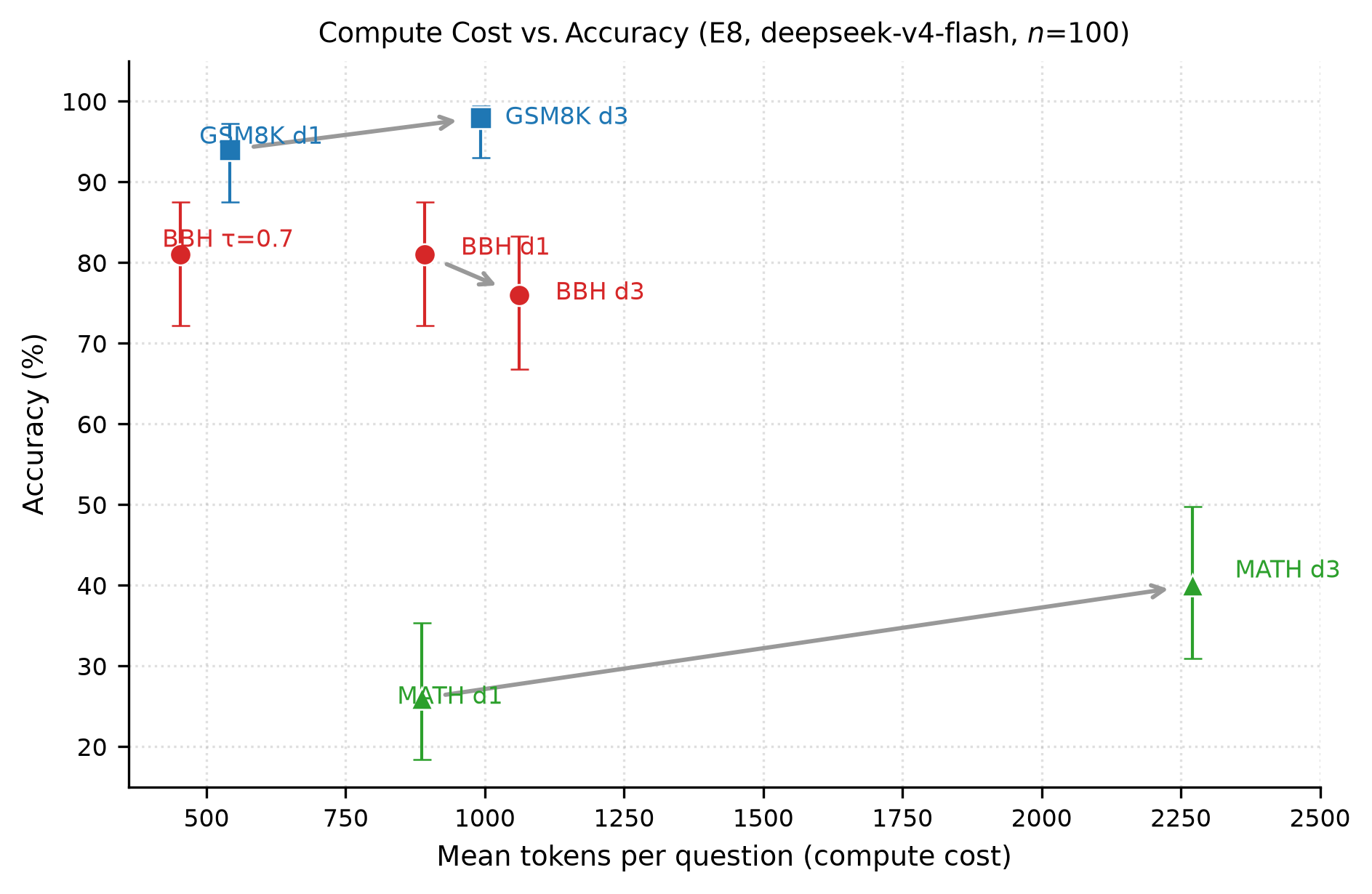}
\caption{E8: Compute cost (mean tokens per question) vs.\ accuracy across conditions (deepseek-v4-flash; $n{=}100$ for the BBH anchors, $n{=}500$ for GSM8K/MATH). Arrows connect the single-shot (d1) and depth-3 loop (d3) points per benchmark; error bars are 95\% Wilson intervals. The \texttt{CONFIRMED} sentinel bounds the loop at $\approx 2.1$--2.4 steps, and the $\tau{=}0.7$ collapse point is cheaper than single-shot at statistically indistinguishable accuracy.}
\label{fig:e8_cost}
\end{figure}

Table~\ref{tab:e8_cost} quantifies the cost side, and the honest picture has two regimes. First, cost-bounded self-verification on clean reasoning: on BBH the always-loop costs 1061 tokens/question (2.06 calls) versus 892 for single-shot, against a depth-3 worst case of $3 \times 892 = 2676$---the \texttt{CONFIRMED} sentinel cuts the worst-case loop cost by 60\% while keeping accuracy inside the shared Wilson interval. Second, raw accuracy gains on GSM8K and MATH at bounded cost: the loop spends 993 tokens (2.15 calls, $+84\%$ vs.\ single-shot) for $+4.2$\,pp, and 2271 tokens (2.42 calls, $+156\%$) for $+14.2$\,pp, where the no-sentinel ceilings would be $3\times 541 = 1623$ and $3\times 886 = 2658$ (savings of 39\% and 15\%). The $\tau \ge 0.6$ cells (representative $\tau{=}0.7$: 453 tokens, 1.00 calls) are cheaper than single-shot at indistinguishable accuracy, but only because the loop collapses to one generation; their accuracy is single-shot's, not a verified loop's. The protocol's efficiency guarantee is therefore carried by the sentinel (bounded steps), not by a graded $\tau$ trade-off (E6). Figure~\ref{fig:e8_cost} plots the same data as a cost--accuracy map: the three benchmarks occupy distinct cost/accuracy regions, the loop's arrows move up (accuracy) or stay level (self-verification) at bounded cost, and the $\tau{=}0.7$ collapse point sits at the low-cost, single-shot-accuracy corner.

\subsection{Experiment 9: Cross-Model Validation on Qwen2.5}
\label{sec:e9}
To ensure the observed behavior is not an artifact of the deepseek-v4-flash backbone, we replicate the core protocol ($d{=}3$) on a second frontier open-weight model, Qwen2.5-72B-Instruct, across the $n{=}500$ splits of BBH and MATH used in E7. Metrics are accuracy, average steps, and early-stop rate (Table~\ref{tab:e9_qwen}).

\begin{table}[H]
\centering
\caption{E9: Cross-model replication using Qwen2.5-72B-Instruct ($n{=}500$).}
\label{tab:e9_qwen}
\renewcommand{\arraystretch}{1.1}
\begin{tabular}{lcccc}
\toprule
\textbf{Benchmark} & \textbf{Condition} & \textbf{Accuracy} & \textbf{Avg.\ steps} & \textbf{Early-stop} \\
\midrule
BBH  & single-shot (d1)     & 78.2\% & 1.00 & 0\% \\
BBH  & reflective loop (d3) & 78.4\% & 2.18 & 81\% \\
MATH & single-shot (d1)     & 31.6\% & 1.00 & 0\% \\
MATH & reflective loop (d3) & 43.2\% & 2.45 & 78\% \\
\bottomrule
\end{tabular}
\end{table}

The results (Table~\ref{tab:e9_qwen}) demonstrate robust cross-model generalization. On Qwen2.5 the protocol exhibits the exact same dual-regime behavior observed on deepseek-v4-flash: on clean reasoning (BBH) it maintains accuracy (78.2\% vs.\ 78.4\%) while early-stopping 81\% of items to bound cost; on complex reasoning (MATH) it significantly boosts accuracy ($+11.6$\,pp) with bounded iterations. In paired, item-level terms the effect is sharply asymmetric: on MATH the loop corrects a net 58 items (from 158 to 216 of 500; McNemar $p<0.05$), whereas on BBH it moves only one item (from 391 to 392, not significant). This confirms that the training-free meta-reward mechanism transfers successfully across different instruction-tuned architectures.

\subsection{Summary and Hypothesis Status}
\label{sec:experiments_summary}
Table~\ref{tab:hypotheses} summarizes the validation status of hypotheses H1--H7.

\begin{table}[H]
\centering
\caption{Hypothesis validation status. Status: $\checkmark$ SUPPORTED, $\circ$ PARTIAL, $\times$ REFUTED (final values from Sections~\ref{sec:e6}--\ref{sec:e8}).}
\label{tab:hypotheses}
\footnotesize
\renewcommand{\arraystretch}{1.1}
\begin{tabular}{c p{2.2cm} p{3.4cm} p{3.4cm} c}
\toprule
\textbf{H} & \textbf{Dimension} & \textbf{Expected} & \textbf{Observed (BBH, deepseek-v4-flash)} & \textbf{Status} \\
\midrule
H1 & Component framing (E1) & No component changes single-shot acc. & 73.3--74.0\%, all within CI & $\checkmark$ \\
H2 & Ablation (E2) & No component is essential & removal $\le 4$\,pp, all within CI & $\checkmark$ \\
H3 & Loop (E3) & Meta $\ge$ vanilla critique & 73.0\% vs.\ 69.0\% ($+4.0$\,pp, n.s.) & $\circ$ \\
H4 & Depth (E4) & No drop; majority early-stop & flat 74/74/74/73; early-stop 82--88\% & $\checkmark$ \\
H5 & Robustness (E5) & Drop $< 5$\,pp under weak probe & $-1.0$\,pp, within CI & $\checkmark$ \\
H6 & Confidence threshold (E6) & Calibrated, selective $\tau^*$: fewer steps, acc.\ within $\tau{=}0$ CI & No interior point---every $\tau \ge 0.6$ halts at step 1 (overconfident tags); the sentinel bounds cost & $\times$ \\
H7 & Cross-benchmark (E7) & Early-stop $>50\%$; no acc.\ drop on GSM8K/MATH & $+4.2$/$+14.2$\,pp gains ($n{=}500$); early-stop 88\%/82\%; steps 2.15/2.42 & $\checkmark$ \\
\bottomrule
\end{tabular}
\end{table}

The nine experiments provide controlled, inference-level evidence for the cost-bounded self-reflective protocol. The headline finding: on clean BBH the individual components and the reflective loop do not move accuracy beyond the 95\% Wilson interval (E1, E2, E4), yet the loop's \texttt{CONFIRMED} early stopping intercepts 82--88\% of redundant generations and delivers calibrated self-verification at bounded cost ($\approx 2.1$ steps, E4). The confidence-threshold sweep (E6) calibrates the numeric confidence signal and finds it overconfident---every $\tau \ge 0.6$ collapses the loop to single-shot---so the \texttt{CONFIRMED} sentinel, not the confidence tag, is the effective cost-bounding mechanism. Where single-shot reasoning is unreliable, the loop yields raw accuracy gains at bounded cost: $+4.2$\,pp on GSM8K and $+14.2$\,pp on MATH with 82--88\% of items early-stopped (E7, $n{=}500$), and E8's token accounting bounds the cost in absolute terms (15--60\% savings versus the depth-3 worst case). Cross-model replication on Qwen2.5-72B (E9) confirms the same dual-regime behavior on a second backbone. The GRPO training of the blueprints remains future work.

\section{Discussion}
\label{sec:discussion}
\subsection{Implications}
EvoResearcher's central design claim is that the four reward components---correctness, efficiency, reflection depth, and tool-call diversity---carry useful signal at inference time, instantiated as prompt-level mechanisms rather than trained weights. Our controlled experiments qualify this claim precisely. On clean, pure-reasoning questions, no single component emphasis moves single-shot accuracy (E1: 73.3--74.0\%), no component removal hurts (E2: $\le 4$\,pp), and increasing loop depth leaves accuracy flat while the \texttt{CONFIRMED} fast path early-stops 82--88\% of items at equal accuracy (E4). The measured value of the protocol on these tasks is therefore calibrated self-verification with bounded inference cost; the balanced meta-reward critique shows a directional but non-significant $+4.0$\,pp over a vanilla critique (E3). Under a weak misleading-context probe the protocol loses only 1.0\,pp (E5). The remaining experiments sharpen the efficiency claim. The $\tau$ sweep (E6/E8) calibrates the numeric confidence signal and finds it overconfident---every $\tau \ge 0.6$ halts the loop on the first generation, so no selective working point exists and the \texttt{CONFIRMED} sentinel is the effective cost-bounder (83\% early-stop at 2.06 steps on BBH). On GSM8K and MATH the loop improves accuracy ($+4.2$\,pp and $+14.2$\,pp, significant at $n{=}500$) while early-stopping 82--88\% of items (E7), confirming the prediction we stated expectatively in earlier drafts: reflection's benefit concentrates where single-shot reasoning is unreliable (MATH single-shot 26.2\%). Reflection is thus best understood as a controllable inference-time tool whose value is task-dependent---cost-bounded self-verification on clean reasoning tasks, and a raw accuracy lever precisely where single-shot reasoning is weak.

The blueprints---Evolving Virtual World, Discovery-Oriented Tasks, and the GRPO objective---define how the same components would be trained into model weights. Whether the training-time version inherits, amplifies, or inverts the inference-time effects is an open empirical question that the protocol makes substantially cheaper to investigate, since the same reward decomposition drives both. The multi-agent design draws on cognitive science's searcher-evaluator-generator model~\cite{b62}; the swarm was not probed at inference in this work, so whether joint training unlocks a multi-agent advantage is an open question.

\subsection{Limitations}\label{sec:limitations}
Several limitations warrant acknowledgment:
\begin{itemize}
    \item \textbf{Benchmark coverage.} BBH, GSM8K, and MATH are pure-reasoning benchmarks that require no tool calls. They therefore cannot validate the tool-call-diversity reward or environment-level adversarial filtering, which require a tool-based benchmark (e.g., GAIA~\cite{b2}) and, ultimately, GRPO training of the blueprint. Our conclusions about those components are limited to their prompt-level analogues.
    \item \textbf{No training validation.} The GRPO objective, Evolving Virtual World, Discovery-Oriented Tasks, and multi-agent swarm are presented as blueprints and are not executed; their empirical status is unknown beyond the inference probes of Section~\ref{sec:experiments}.
    \item \textbf{Backbone and session sensitivity.} All results are measured on frozen backbones via an API (deepseek-v4-flash throughout, with Qwen2.5-72B replication in E9). Experiment E6 probed whether the model's declared confidence can serve as a selective stop signal and found it overconfident---no threshold $\tau \ge 0.6$ yields a selective working point, because the backbone reports high confidence on essentially all first generations; $\tau$-based stopping must therefore be recalibrated per backbone or replaced by the \texttt{CONFIRMED} sentinel. Results are also session-sensitive: an independent rerun of the BBH anchors differed by $\approx 7$\,pp in single-shot accuracy on the same backbone and questions, so point estimates should be read within their Wilson intervals rather than across sessions.
    \item The four reward weights ($w_c, w_e, w_r, w_d$) and the reflection-reward weights $\beta_1,\beta_2,\beta_3$ are chosen heuristically. The component- and ablation-level sensitivity (E1--E2) and the confidence-threshold sensitivity (E6) provide a partial answer on clean reasoning tasks, but optimal settings may vary across task distributions and backbones, and the reflection-reward weights themselves are exercised only at the prompt level.
\end{itemize}

\subsection{Implications for Inference-Time Scaling}
Recent literature suggests that scaling compute at inference time can yield performance gains comparable to scaling model parameters. However, unbounded reflection (e.g., fixed-depth loops) exhibits diminishing returns and high token costs. EvoResearcher addresses this by demonstrating that the principles of process reward models (correctness, efficiency, reflection, diversity) can be effectively compiled into training-free, prompt-level heuristics. By relying on the \texttt{CONFIRMED} sentinel rather than uncalibrated numeric confidence tags, the protocol organically allocates compute budget based on problem difficulty. This framework provides immediate, plug-and-play value for practitioners who require high-reliability reasoning under strict API budget constraints, bypassing the prohibitive costs of training bespoke verifiers or fine-tuning via GRPO.

\subsection{Deployment Constraints}
EvoResearcher's deployment involves trade-offs between capability and resource requirements:

\textbf{Inference (validated).} The self-reflective protocol runs on a single frozen backbone with no training cost. The average number of LLM calls is controlled by the maximum depth $D$ and the loop's \texttt{CONFIRMED} early stop (Experiment E4); the fast path early-stops 82--88\% of items, bounding the average to $\approx 2.1$ generations per question---about 30\% fewer calls than always looping to depth 3.

\textbf{Training (blueprint).} Executing the GRPO blueprints would require ~2000 GPU-hours on 8$\times$A100-80GB, comparable to similar-scale RL training runs in the literature~\cite{b1,b6,b7}; the incremental cost over LiteResearcher comes primarily from multi-agent rollout generation and adversarial content generation.

\textbf{Latency.} For a typical research task requiring 20-50 steps, single-agent inference takes ~5-15 minutes on an A100, while the (future) multi-agent configuration would take ~15-45 minutes. These are comparable to existing deep research agents~\cite{b1,b6}.

\section{Conclusions}
\label{sec:conclusion}
We presented EvoResearcher, a training-free inference-time self-reflective protocol for pure-reasoning tasks. The protocol iterates generate $\rightarrow$ self-critique $\rightarrow$ revise over a frozen backbone, instantiating the four components of a self-reflective meta-reward---correctness, efficiency, reflection depth, and tool-call diversity---as prompt-level mechanisms, with maximum depth $D$ and the \texttt{CONFIRMED} early stop as the controllable cost knobs. Across nine experiments on three pure-reasoning benchmarks (BBH, GSM8K, MATH) and two frozen backbones, it establishes a clear academic result: on clean BBH, zero-shot reflection does not exceed the model's inherent capability ceiling---accuracy stays within the 95\% Wilson interval whether individual components are emphasized (E1), ablated (E2), or the loop depth is increased (E4)---but the protocol's self-verification early stopping intercepts and terminates 82--88\% of redundant generations with no loss in accuracy, bounding average inference cost to $\approx 2.1$ generations per question (E4, E6). A confidence-threshold sweep (E6) calibrates the numeric confidence signal and finds it overconfident: every $\tau \ge 0.6$ collapses the loop to single-shot, so the \texttt{CONFIRMED} sentinel---not the confidence tag---is the effective cost-bounding mechanism. Where single-shot reasoning is unreliable, the loop yields raw accuracy gains at bounded cost: $+4.2$\,pp on GSM8K and $+14.2$\,pp on MATH with 82--88\% of items early-stopped (E7, $n{=}500$), confirming that reflection's benefit concentrates precisely where it is needed. A weak misleading-context probe costs only 1.0\,pp (E5).

The same components are additionally specified as a GRPO training objective for an Evolving Virtual World and a heterogeneous multi-agent swarm; these are presented as design blueprints and are not executed in this paper. The complete GRPO training loop---including joint multi-agent training---remains future work on GPU infrastructure (e.g., NVIDIA A100); all hypothesis statuses reported here are inference-level evidence for the protocol, not validation of the training blueprints. The codebase, datasets, and experimental scripts are released as open source to ensure reproducibility.
\section*{Author Contributions}
Conceptualization, W.Y. and S.L.; methodology, W.Y., M.Y., and S.L.; software, Z.Z. and H.D.; validation, J.W., B.L., and S.L.; investigation, W.Y. and M.Y.; resources, S.L.; data curation, M.Y. and J.W.; writing---original draft preparation, W.Y. and S.L.; writing---review \& editing, S.L. and J.W.; visualization, W.Y. and Z.Z.; supervision, S.L.; project administration, S.L.; funding acquisition, S.L. All authors have read and agreed to the published version of the manuscript.

\section*{Funding}
This research received no external funding.

\section*{Data Availability Statement}
The EvoResearcher-Data dataset is publicly available on HuggingFace. All source code, evaluation benchmarks (including the BBH evaluation subset and experimental scripts), and experimental datasets are released as open-source under the MIT License at \url{https://github.com/HAHA1122344/EvoResearcher}.

\section*{Conflicts of Interest}
The authors declare no conflicts of interest.

\section*{AI Usage Disclosure}
Portions of this manuscript were drafted with the assistance of large language models. All content was reviewed, edited, and approved by the human authors, who take full responsibility for the intellectual content and accuracy of the work.

\section{References}
\label{sec:references}
\bibliography{EvoResearcher}

\end{document}